%% file: main.tex
\documentclass[conference]{IEEEtran}
\IEEEoverridecommandlockouts
\usepackage{cite}
\usepackage{amsmath,amssymb,amsfonts}
\usepackage{algorithm}
\usepackage{algpseudocode}
\usepackage{booktabs}
\usepackage{multirow}

\algrenewcommand\algorithmicfor{\textbf{for}}
\algrenewcommand\algorithmicdo{\textbf{do}}
\algrenewcommand\algorithmicif{\textbf{if}}
\algrenewcommand\algorithmicthen{\textbf{then}}
\algrenewcommand\algorithmicend{\textbf{end}}

\algrenewcommand\algorithmicrequire{\textbf{input:}}
\algrenewcommand\algorithmicensure{\textbf{output:}}

\usepackage{graphicx}
\usepackage{textcomp}
\usepackage{xcolor}
\def\BibTeX{{\rm B\kern-.05em{\sc i\kern-.025em b}\kern-.08em
    T\kern-.1667em\lower.7ex\hbox{E}\kern-.125emX}}

\input{macros}

\begin{document}


\title{\LARGE \textbf{
Scenario MPC with STL Specifications and Pareto-Based Feasibility Repair
}\\
{\large }


\author{\large Tianhao Wu$^1$, Yiwei Lyu$^2$}
\thanks{ 
 $^1$ The author is with the Department of Computer Science, University of Southern California. Email: wutianha@usc.edu
 
 $^2$ The author is with the Department of Computer Science and Engineering, Texas A\&M University. Email: yiweilyu@tamu.edu
 }
}

\maketitle


\begin{abstract}
Temporal logic is a formal language for reasoning about system behaviors over time. Signal temporal logic (STL), in particular, has been used to encode spatio-temporal requirements for control synthesis in multi-agent systems, often under the assumption that agents are cooperative and their dynamics are known. However, real-world multi-agent applications, such as autonomous driving, typically involve stochastic and uncontrollable agents. Recent work explored robust control with worst-case or probabilistic formulations, but remains limited in that it either (1) certifies strict satisfaction of STL constraints without addressing feasibility recovery, or (2) relaxes infeasible constraints with ego-centric objectives. In this paper, we propose a model predictive control (MPC) framework that treats feasibility repair as a Pareto optimization problem to explicitly characterize tradeoffs among agent objectives. We further provide a probabilistic certificate on STL violation rate to formally quantify uncertainty under stochastic and uncontrollable agents. The proposed framework is evaluated on two autonomous driving scenarios. Results show that the framework recovers feasible control with demonstrated safe behaviors.

\end{abstract}

\input{revision/introduction}

\input{revision/preliminaries}
\input{revision/method}
\input{revision/experiments}
\input{revision/conclusion}

\bibliographystyle{IEEEtran}
\bibliography{refs}

\end{document}

%% file: macros.tex
\usepackage{amsmath, amssymb, amsthm}
\usepackage{mathtools}          
\usepackage{bm}                 
\usepackage{xspace}             
\usepackage{xcolor}



%% file: revision/introduction.tex
\section{Introduction}

Temporal logic (TL) is a class of formal specification languages that describes how systems should behave over time \cite{rescher2012temporal}. With this capability of expressing diverse spatio-temporal requirements, logical formalisms such as Linear Temporal Logic (LTL) \cite{pnueli1977temporal} and Metric Temporal Logic (MTL) \cite{koymans1990specifying} are widely used to encode high-level robot tasks and enable systematic verification \cite{kress2009temporal, kloetzer2008fully, karaman2008vehicle}. 
A central appeal of temporal logic is that it connects symbolic requirements with control synthesis for robotic systems. In automaton-based pipelines, a formula is compiled into a finite-state automaton that tracks progress toward satisfaction \cite{baier2008principles}. In optimization-based frameworks, a formula is translated into constraints and objectives while jointly optimized with performance criteria such as efficiency and smoothness \cite{gautam2025rrt, ren2024ltl}.

In particular, Signal Temporal Logic (STL) defines a quantitative semantics, which assigns a real-valued robustness score indicating by how much a specification is satisfied \cite{donze2010robust}. This quantitative measure of satisfaction makes STL attractive for optimization-based control, where robustness can be maximized as objectives or enforced as task constraints \cite{yu2023model, sadraddini2015robust, yu2026signal}. 
A mixed-integer linear programming (MILP) encoding was introduced to translate STL specifications into constraints while preserving their Boolean and temporal semantics \cite{raman2014}.

However, feasibility of these STL-constrained optimizations is not always guaranteed. For example, requiring a robot to reach a location beyond its physical limitations within some time produces an infeasible optimization problem when the corresponding STL specification is enforced with the robot dynamics. Existing methods recover feasibility through spatial relaxation \cite{ghosh2016diagnosis,charitidou2021barrier}, temporal relaxation \cite{buyukkocak2025resilient,verhagen2023}, or partial satisfaction \cite{cardona2022partial}. Spatial relaxation introduces slack variables to quantify predicate-level violations, whereas temporal and partial relaxation modify the temporal interval or allow selected subformulae to remain unsatisfied. Although bounded temporal relaxation limits the allowed adjustment using a user-defined tolerance \cite{buyukkocak2022}, disabling a specification, even just temporarily or partially, may be undesirable when continuous satisfaction is needed. Hence, in this paper, we use bounded spatial relaxation, which maintains satisfaction during the specified interval while explicitly limiting its deviation from the original specification.

STL-based control has received much attention in multi-agent systems \cite{buyukkoccak2021distributed,lindemann2019control,sun2022multi,pant2018fly,meng2023signal}. However, existing formulations often assume that agents are cooperative and that their dynamics are known a priori, which are hard to satisfy in real-world systems.  \cite{yu2026signal} considers STL satisfaction in the presence of uncontrollable agents with unknown intentions, but does not address feasibility recovery when the initial problem is infeasible. 
Moreover, established approaches often consider ego-centric objectives in STL-constrained optimizations. For example, \cite{cui2025stackelberg} formulates a leader-follower game in which the ego agent and uncontrollable agents optimize over their separate objectives at each step. Other work consider robustness maximization \cite{raman2014} or relaxation minimization \cite{buyukkocak2025resilient}, which still focuses on specification satisfaction of the ego agent. Such formulations produce ego-optimal yet collectively suboptimal solutions without taking into account the performance of other agents.
\cite{schon2026spatiotemporal} applies multi-objective reasoning but addresses tradeoffs between STL spatial and temporal robustness instead of agent-specific objectives. 


Another challenge is uncertainty quantification under stochastic agents constrained by STL specifications. These agents have random future behaviors that require STL satisfaction to be evaluated over sampled realizations or probabilistic constraints. Previous work has addressed such uncertainty using worst-case formulations \cite{farahani2015robust} and chance-constrained methods \cite{farahani2018shrinking}. However, worst-case satisfaction over finitely many samples does not quantify the violation probability under unseen realizations, while chance-constrained methods may require prior information about the disturbance distribution.

For these reasons, fundamental challenges exist in establishing guarantees from sampling-based STL satisfaction under uncertainty, and in the case of infeasibility, recovering a feasible control that considers the performance of all agents. To the best of our knowledge, this is the first work that jointly considers feasibility recovery under uncertainty and infeasible STL constraints, while addressing multi-objective reasoning to characterize tradeoffs among agent-specific objectives. In this paper, we use a sampled worst-case model predictive control (MPC) formulation where the STL specifications are encoded as MILP constraints. We further derive a probabilistic certificate to formally quantify STL violation rate under uncertainty induced by uncontrollable agents, without requiring an explicit distributional model. Our \textbf{main contributions} are:
\begin{itemize}
    \item We provide probabilistic verification that quantifies STL violation rate in a worst-case formulation under stochastic and uncontrollable agents.
    
    \item We formulate feasibility recovery as a Pareto optimization problem that directly characterizes tradeoffs among agent-specific objectives.

    \item We show effectiveness of our proposed framework in two autonomous driving case studies with demonstrated safe behaviors.
\end{itemize}

%% file: revision/preliminaries.tex
\section{Preliminaries}


\subsection{Signal Temporal Logic}

A signal $S(t)$ is a function $S:\tau \rightarrow \mathbb{R}^n$, where $\tau \subseteq \mathbb{R}_{\geq 0}$ is a time domain. An STL formula $\varphi$ is recursively defined as:
\begin{equation}
\varphi ::= \top \mid \mu \mid \neg \varphi \mid
\varphi_1 \wedge \varphi_2 \mid
F_I \varphi \mid
G_I \varphi.
\end{equation}
where $\top$ is boolean true.
$\mu := l(S(t)) \ge 0$ is an atomic predicate with predicate function $l : \mathbb{R}^n \to \mathbb{R}$.
$\neg$ (negation) and $\wedge$ (conjunction) are boolean logical operators.
$F$ (eventually) and $G$ (always) are temporal operators defined over time interval $I=[a, b] \subseteq \tau$.

The STL \textit{qualitative (Boolean)} semantics \cite{maler2004} determines whether a formula $\varphi$ is satisfied by signal $S$ at time $t$, denoted as $(S,t) \models \varphi$. 
Its \textit{quantitative (robustness)} semantics \cite{donze2010robust} assigns a real-valued robustness score that quantifies the amount of satisfaction or violation. Denoted as $\rho(\varphi,S,t)$, the robustness is recursively defined as:
\begin{equation}
\begin{aligned}
\rho(\mu,S,t) &= l(S(t)), \\
\rho(\neg \varphi,S,t) &= -\rho(\varphi,S,t), \\
\rho(\varphi_1 \wedge \varphi_2,S,t) 
&= \min\!\big(\rho(\varphi_1,S,t), \rho(\varphi_2,S,t)\big), \\
\rho(\varphi_1 \vee \varphi_2,S,t) 
&= \max\!\big(\rho(\varphi_1,S,t), \rho(\varphi_2,S,t)\big), \\
\rho(G_I \varphi,S,t) 
&= \inf_{t' \in t+I} \rho(\varphi,S,t'), \\
\rho(F_I \varphi,S,t) 
&= \sup_{t' \in t+I} \rho(\varphi,S,t').
\end{aligned}
\end{equation}
Meanwhile, robustness is \textit{sound} and \textit{complete} with respect to its Boolean semantics, which allows boolean satisfaction to be determined from the sign of a robustness score.
\begin{equation}
\rho(\varphi,S,t) \ge 0 \Leftrightarrow (S,t) \models \varphi.
\label{eq:stl_robustness}
\end{equation}




\subsection{Pareto Optimization} 

Also known as multi-objective optimization, Pareto Optimization solves multiple objective functions subject to shared constraints. Consider the problem,

\begin{equation}
\begin{aligned}
\min_{z \in \mathcal{Z}} \quad 
& \mathbf{g}(z) = \big( g_1(z), \dots, g_n(z) \big)^T \\
\text{s.t.} \quad 
& h_j(z) \le 0, \quad j = 1, \dots, m \\
& l_k(z) = 0, \quad k = 1, \dots, e
\end{aligned}
\label{eq:moo}
\end{equation}
where $z$ is the decision variable, 
$g_1(z), \dots, g_n(z)$ are objective functions, and $h_j(z), l_k(z)$ are constraints of the optimization.
Generally, no single solution can optimize all objectives simultaneously. To formally characterize this tradeoff among objectives, we introduce the following concepts.



\textit{\textbf{Definition 1} (Weak Pareto Optimality):}
A point $z^\star \in \mathcal{Z}$ is weakly Pareto optimal if there exists no other feasible point that strictly improves all objectives:
\begin{equation}
\label{eq:weak_pareto}
\nexists\, z \in \mathcal{Z} \text{ s.t. }
\forall k \in \{1,\ldots,n\}, \quad
g_k(z) < g_k(z^\star).
\end{equation}

\textit{\textbf{Definition 2} (Pareto Optimality):}
A point $z^\star \in \mathcal{Z}$ is Pareto optimal if it is not dominated by any other feasible point.
\begin{equation}
\nexists\, z \in \mathcal{Z} \text{ s.t. }
\begin{cases}
\forall k \in [1, n], \quad g_k(z) \le g_k(z^\star), \\
\exists k \in [1, n], \quad g_k(z) < g_k(z^\star).
\end{cases}
\end{equation}


\textit{\textbf{Definition 3} (Pareto front):}
Let $\mathcal{P}$ be a set that contains all Pareto optimal points. The image of $\mathcal{P}$ in the objective space is called the Pareto front $\mathcal{F}$, where
\begin{equation}
\mathcal{F}
=
\left\{
\big(g_1(z),\dots,g_n(z)\big)
\;\middle|\;
z \in \mathcal{P}
\right\}.
\end{equation}

%% file: revision/method.tex
\section{Method}

\subsection{Problem Statement}

In this work, we consider control synthesis from STL specifications in multi-agent systems with stochastic and uncontrollable agents. We are inspired by two fundamental research questions: \emph{(1) How can we quantify uncertainty when an agent performs well on sampled interactions?} \emph{(2) How to consider the overall performance of all agents rather than ego-centric objectives in our control design?} 
These questions motivate two complementary technical objectives that address out-of-sample reliability and agent-aware feasibility recovery.
In this paper, our objectives are to:
\begin{enumerate}
    \item \textbf{Assess strict feasibility:} we determine whether a control that satisfies the sampled interactions has an acceptable risk of violating the STL specifications under an unseen realization.
    
    \item \textbf{Recover feasible control:} when strict STL satisfaction is infeasible, we identify physically meaningful relaxations and control decisions that balance their consequences for the other agents.
\end{enumerate}


\subsection{System Setting}
Consider a robot with discrete-time dynamics of the form
\begin{equation}
x_{t+1} = f(x_t, u_t)
\label{eq:discrete_dynamics}
\end{equation}
where $x_t \in \mathcal{X} \subseteq \mathbb{R}^{n_x}$ and $u_t \in \mathcal{U} \subseteq \mathbb{R}^{n_u}$ are the system state and control input at time $t$. Here, $\mathcal{X}$ and $\mathcal{U}$ denote the state space and set of admissible control inputs. Given an initial state $x_0$, a time horizon $T$, and a control sequence $\textbf{u}_T = u_0u_1...u_{T-1}$, the resulting trajectory $\textbf{x}_T = x_0x_1...x_T$ is obtained by applying control $\textbf{u}_T$ from state $x_0$. In model predictive control, only the first control $\mathbf{u}_0^\star$ from the optimal control sequence $\mathbf{u}_T^\star$ is applied at each time step. Subsequently, the prediction horizon is advanced by one step, and the optimization is re-solved using the latest information.

The robot in \eqref{eq:discrete_dynamics} operates around $l$ uncontrollable agents with unknown dynamics and intentions. Let $\hat{\mathbf{x}}^i_T=(\hat{x}^i_0,\ldots,\hat{x}^i_T)$ be a sample trajectory of agent $i\in\{1,\ldots,l\}$, with $\omega=\left(\hat{\mathbf{x}}^1_T,\ldots,\hat{\mathbf{x}}^l_T\right)$ representing a joint sample over all $l$ agents. Here, we assume access to sample trajectories of uncontrollable agents, which can be obtained from a probabilistic trajectory predictor \cite{Ivanovic_2019_ICCV, chai2019multipath} or generated by applying Gaussian noise\cite{wu2022forecasting}.

\subsection{Scenario-Based Worst-Case MPC}
\label{sec:worst-case-mpc}

We next introduce our sampled worst-case MPC formulation with STL constraints. Given a set of STL specifications denoted by $\Phi=\{\varphi_1,\ldots,\varphi_K\}$, the problem is formally stated as follows.

\textit{\textbf{Problem:}}
Given an initial state $x_0$, a time horizon $T$, a cost function $J$, a set of STLs $\Phi$, and $N$ joint trajectory samples
$\Omega=\{\omega^{(1)},\ldots,\omega^{(N)}\}$, we solve
\begin{equation}
\begin{aligned}
\underset{\mathbf{u}_T}{\operatorname{argmin}}
\quad &
J\bigl(\mathbf{x}_T,\mathbf{u}_T\bigr)
\\
\text{s.t.}\quad
&
\rho\bigl(\varphi_k,\mathbf{x}_T,\omega^{(n)}\bigr)
\geq 0,
\quad \forall\varphi_k\in\Phi, ~\forall\omega^{(n)}\in\Omega
\\
&
x_{t+1}=f(x_t,u_t),
\\
&
x_t\in\mathcal{X}, \quad
u_t\in\mathcal{U}.
\end{aligned}
\label{eq:strict_mpc_stl}
\end{equation}
where $\rho\bigl(\varphi_k,\mathbf{x}_T,\omega^{(n)}\bigr)\geq 0$, by equivalence \eqref{eq:stl_robustness}, enforces satisfaction of specification $\varphi_k$ on sample $\omega^{(n)}$.


\subsection{Probabilistic Certificate under Uncertainty}
\label{sec:certificate}

While the sampling-based formulation provides an efficient way to check whether there is a foreseeable violation, satisfying STL constraints over some finite samples does not establish reliability under unseen realizations, for example, if the samples are not representative of the underlying distribution.
In this section, we derive a probabilistic certificate for the STL violation rate with an independent validation procedure. 

If \eqref{eq:strict_mpc_stl} is feasible, it produces an optimal control sequence $\mathbf{u}_T^\star$ and a corresponding trajectory $\mathbf{x}_T^\star$, which satisfy all $K$ STL specifications over $N$ sampled joint trajectories used in the optimization. We want to determine whether this feasibility over finite samples also implies a low STL violation rate for previously unseen realizations of agent trajectories. Let $\Omega_N=\{\omega^{(1)},\ldots,\omega^{(N)}\}$ denote the optimization samples, assumed to be independently and identically distributed according to an unknown distribution $P$, i.e., $\omega^{(n)}\overset{\mathrm{i.i.d.}}{\sim}P$. Note that explicit knowledge of $P$ is not required.

To evaluate the fixed solution $(\mathbf{u}_T^\star,\mathbf{x}_T^\star)$ independent of the optimization samples, we draw $M$ additional joint trajectories from the same distribution to form the validation set $\widetilde{\Omega}_M=\{\tilde{\omega}^{(1)},\ldots,\tilde{\omega}^{(M)}\}$. We say that $(\mathbf{u}_T^\star,\mathbf{x}_T^\star)$ passes validation if it satisfies every STL specification over each trajectory sample,
\begin{equation}
\rho\bigl(\varphi_k,\mathbf{x}_T^\star,\tilde{\omega}^{(m)}\bigr) \geq 0,
\qquad
\forall\varphi_k\in\Phi,\quad
\forall\tilde{\omega}^{(m)}\in\widetilde{\Omega}_M .
\label{eq:validation}
\end{equation}
We define the true STL violation probability of $\mathbf{u}_T^\star$ as,
\begin{equation}
p_{\mathrm{viol}}(\mathbf{u}_T^\star)
=
\Pr_{\omega\sim P}
\left[
\exists \varphi_k\in\Phi:
\rho\bigl(\varphi_k,\mathbf{x}_T^\star,\omega\bigr)<0
\right].
\label{eq:violation_probability}
\end{equation}
which denotes the probability that an unseen joint trajectory causes at least one STL specification to be violated. The following theorem assumes successful validation to provide a probabilistic upper bound on $p_{\mathrm{viol}}(\mathbf{u}_T^\star)$.

\textit{\textbf{Theorem 1} (Probabilistic certificate):}
Let $\mathbf{u}_T^\star$ be a solution of \eqref{eq:strict_mpc_stl} obtained using $N$ i.i.d.\ samples $\Omega_N$. Let $\widetilde{\Omega}_M$ be a validation set of $M$ i.i.d.\ samples independently drawn from the same distribution $P$. If $\mathbf{u}_T^\star$ passes validation \eqref{eq:validation}, then with confidence at least $1-\beta$,
\begin{equation}
p_{\mathrm{viol}}(\mathbf{u}_T^\star)\leq\epsilon,
\label{eq:probabilistic_certificate}
\end{equation}
provided
\begin{equation}
M \geq
\left\lceil
\frac{\log(1/\beta)}
{-\log(1-\epsilon)}
\right\rceil
\label{eq:validation_sample_complexity}
\end{equation}


\textit{\textbf{Proof.}}
Conditioned on the $N$ training scenarios, the solution $\mathbf{u}_T^\star$ is fixed and independent of the validation set $\widetilde{\Omega}_M$. Let $p := p_{\mathrm{viol}}(\mathbf{u}_T^\star)$, then $\mathbf{u}_T^\star$ satisfies all STL specifications on a single validation sample with probability $1-p$. Given the validation set contains $M$ i.i.d. samples, we know
\begin{equation}
\Pr\!\left(
\rho\bigl(\varphi_k,\mathbf{x}_T^\star,\tilde{\omega}^{(m)}\bigr)\geq 0,\;
\forall k,m
\,\middle|\,
\mathbf{u}_T^\star
\right)
=
(1-p)^M   
\end{equation}
by Lemma 1,
\begin{equation}
\mathbf{1}\{p>\epsilon\}(1-p)^{M}\leq\mathbf{1}\{p>\epsilon\}(1-\epsilon)^{M}\leq\beta
\label{eq:lemma1_ineq}
\end{equation}
take expectation on inequalities \eqref{eq:lemma1_ineq}
\footnote{For any event $A$, the expectation of its indicator equals its probability, i.e., $\mathbb{E}[\mathbf{1}\{A\}]=\Pr(A)$.}, we get
\begin{equation}
\Pr_{\Omega,\widetilde{\Omega}_M}\Bigl[\eqref{eq:validation}\text{ holds}\;\wedge\;
p>\epsilon\Bigr]\leq\beta
\end{equation}
or equivalently, validation \eqref{eq:validation} certifies $p_{\mathrm{viol}}(\mathbf{u}_T^\star)\leq\epsilon$ with
a confidence level of at least $1-\beta$.
\hfill$\square$

Thus, the theorem converts successful finite-sample validation into an out-of-sample risk certificate, which addresses our objective of assessing strict feasibility under uncertain agent behavior.
The remaining question is how many validation samples are required to attain a desired violation threshold $\epsilon$ and confidence level $1-\beta$. The following lemma provides this sample-size condition.

\textit{\textbf{Lemma 1} (Validation sample size):}
Let $\epsilon,\beta\in(0,1)$ and $M\in\mathbb{N}$. Then
\begin{equation}
M\geq
\frac{\log(1/\beta)}
{-\log(1-\epsilon)}
\quad\Longleftrightarrow\quad
(1-\epsilon)^M\leq\beta
\label{eq:sample_size}
\end{equation}


\textit{\textbf{Proof.}}
Since $\epsilon\in(0,1)$, $0<1-\epsilon<1$, thus $c:=\log(1-\epsilon)<0$; since $\beta\in(0,1)$, $\log(1/\beta)=-\log\beta>0$. Multiplying the left inequality of \eqref{eq:sample_size} by $c$, we get $M\log(1-\epsilon)=\log\bigl((1-\epsilon)^M\bigr)\leq-\log(1/\beta)=\log\beta$. Hence, $(1-\epsilon)^M\leq\beta$. Since each step is reversible, equivalence \eqref{eq:sample_size} holds.
\hfill$\square$


\textit{\textbf{Remark 1.}}
Validating $\mathbf{u}_T^\star$ requires only evaluating its STL robustness over additional samples, which does not increase the computational complexity of the original optimization.



\textit{\textbf{Remark 2.}}
If $\mathbf{u}_T^\star$ fails the validation, the desired $(\epsilon,\beta)$-certificate cannot be established, but this does not imply the true violation probability exceeds $\epsilon$.

\subsection{Feasibility Repair as Pareto Optimization}

We previously showed a probabilistic certificate when optimization \eqref{eq:strict_mpc_stl} is feasible. However, in a stochastic multi-agent system, we may observe new agents or events that introduce infeasible STL constraints. In this case, the controller needs to recover a solution that partially relaxes the original safety requirements while considering the consequences of the relaxation for other agents. A key limitation of ego-centric feasibility repair is that minimizing only the ego cost or total specification relaxation does not distinguish how a repaired control affects different agents. Use autonomous driving as an example, a hard brake may reduce the ego vehicle's immediate collision risk while imposing substantial risk on a closely following vehicle. Selecting such a repair can therefore produce behavior that is ego-optimal but collectively suboptimal. 

In this section, to explicitly represent these effects, we associate each uncontrollable agent $i\in\{1,\ldots,l\}$ with a cost function $J_i(\mathbf{x}_T,\mathbf{x}_T^i)$ that depends on the ego trajectory and the trajectory of agent $i$, and assume that each cost can be encoded using MILP constraints. We define the collective objective vector
$
\mathbf{J}(\mathbf{x}_T)
=
\bigl(
J_1(\mathbf{x}_T,\mathbf{x}_T^1),\ldots,
J_l(\mathbf{x}_T,\mathbf{x}_T^l)
\bigr)^{T}.
$
Rather than optimizing a single ego-centric objective, we jointly reason over all agent-specific objectives and formulate feasibility repair as the following Pareto optimization problem.

\begin{equation}
\begin{aligned}
\underset{\mathbf{u}_T,\boldsymbol{\xi}}{\operatorname{argmin}}
\quad &
\mathbf{J}(x_T)\\
\text{s.t.}\quad
&
\rho\bigl(\varphi_k,\mathbf{x}_T,\omega^{(n)}\bigr)
\geq -\xi_k,
\quad \forall\varphi_k\in\Phi, ~\forall\omega^{(n)}\in\Omega
\\
&
0\leq\xi_k\leq\bar{\xi}_k,
\\
&
x_{t+1}=f(x_t,u_t),
\\
&
x_t\in\mathcal{X}, \quad
u_t\in\mathcal{U}.
\end{aligned}
\label{eq:pareto_mpc_stl}
\end{equation}
Each specification $\varphi_k$ is relaxed by a non-negative slack variable $\xi_k$, given by $\rho(\varphi_k,\mathbf{x}_T,\omega^{(n)})\geq-\xi_k$, where $\xi_k$ quantifies the amount of violation allowed and is bounded between $0\leq\xi_k\leq\bar{\xi}_k$. Here, $\bar{\xi}_k$ denotes the maximum possible relaxation of $\varphi_k$.
For example, consider $\varphi_k = \Box_{I} dist(\mathbf{x}_T, \mathbf{x}_T^{k}) \geq 2$ that requires a minimum distance of 2 between ego and agent $k$. If we set $\bar{\xi}_k = 1$, then $0\leq\xi_k\leq1$, thus its robustness $\rho(\varphi_k)\geq 2-\xi_k$ is bounded between $[1, 2]$. This allows specification violation up to some defined upper bounds that maintains its semantic meaning. Solving the Pareto optimization returns an optimal solution in the sense that no agent can be strictly better off without causing another agent to be worse. 




\subsection{An $\epsilon$-Constraint Approach}

Common methods for solving Pareto optimization problems include weighted-sum scalarization \cite{zadeh1963optimality}, lexicographic optimization \cite{isermann1982linear}, and evolutionary methods \cite{deb2002fast}. In our setting, the $\epsilon$-constraint method \cite{haimes1971bicriterion} is particularly favorable since it keeps the mixed-integer linear structure of our previous MPC-STL formulation. Specifically, we select one agent
cost function $J_p$ as the primary objective and impose upper bounds $\epsilon_i$ on
the remaining objectives $J_i$ for $i\neq p$, effectively turning them into linear constraints. By varying these $\epsilon_i$ bounds, the method can explore tradeoffs without requiring \emph{a priori} weights or agent priorities necessary for weighted-sum approaches. Furthermore, when each subproblem is solved globally optimal, we obtain a solution that is weakly Pareto optimal \cite{mesquita2023new}. Consider a primary objective $J_p$, $p\in\{1,\ldots,l\}$ and upper bounds
$\epsilon_i$ for all $i\neq p$, an $\epsilon$-constraint subproblem is framed as follows.



\begin{equation}
\begin{aligned}
\underset{\mathbf{u}_T,\boldsymbol{\xi}}{\operatorname{argmin}}
\quad &
J_p(\mathbf{x}_T, \mathbf{x}_T^p)
\\
\text{s.t.}\quad
&
\rho\bigl(\varphi_k,\mathbf{x}_T,\omega^{(n)}\bigr)
\geq -\xi_k,
\quad \forall\varphi_k\in\Phi, ~\forall\omega^{(n)}\in\Omega
\\
&
0\leq\xi_k\leq\bar{\xi}_k,
\\
&
J_i(\mathbf{x}_T,\mathbf{x}_T^i)\leq\varepsilon_i,
\quad \forall i\neq p.
\\
&
x_{t+1}=f(x_t,u_t),
\\
&
x_t\in\mathcal{X}, \quad
u_t\in\mathcal{U}.
\end{aligned}
\label{eq:epsilon_constraint_mpc_stl}
\end{equation}

Since each subproblem is defined with a set of $\epsilon_i$ values, we construct a uniform $\epsilon$-grid $E_i$ for each objective $J_i$, assuming no prior knowledge about which sets of values result in a feasible subproblem. This construction requires prior knowledge or informed estimates of the attainable range $[a_i,b_i]$ for each objective. Given an objective range $[a_i,b_i]$ and grid density $d$, we divide the range uniformly into $d$ intervals,
\begin{equation}
E_i=\left\{a_i+\frac{r}{d}(b_i-a_i)\mid r=1,\ldots,d\right\}.
\end{equation}
For a primary objective $J_p$, the $\epsilon$-constraint problems are solved over all $\varepsilon_i$ combinations in the Cartesian product $\prod_{i\neq p}E_i$.  With $l$ objectives, each choice of primary objective creates $d^{l-1}$ subproblems, resulting in $l\cdot d^{l-1}$ solves in total and a grid-search complexity of $O(l\cdot d^{l-1})$.

Under exact global solution of an \(\epsilon\)-constraint subproblem, the returned optimizer is weakly Pareto optimal under the standard conditions of the \(\epsilon\)-constraint method. In practice, however, our finite grid explores only a subset of possible objective bounds. Consequently, the filtered candidate set should be interpreted as a discrete approximation of the Pareto front. Within this approximation set, we select a final solution of the smallest deviation from nominal control to avoid assigning preferences to agents. 


\begin{algorithm}[t]
\caption{Robust MPC-STL via Pareto-based repair}
\label{alg:pareto_epsilon}
\begin{algorithmic}[1]
\Require Optimization \eqref{eq:strict_mpc_stl}, $(\varepsilon, \beta)$-certificate, objectives $\mathbf{J}(x_T)$, grids $\{E_1,\ldots,E_l\}$

\Ensure Control sequence $\mathbf{u}_T^\star$ 

\State Solve optimization \eqref{eq:strict_mpc_stl}
\If{feasible}
    \State Validate and \Return $\mathbf{u}_T^\star$
\EndIf

\State Candidate solutions $\mathcal{C}\leftarrow\emptyset$
\For{$p=1,\ldots,l$}
    \For{each $\boldsymbol{\epsilon}
        \in\prod_{i\neq p}E_i$}
        \State Solve $\epsilon$-constraint subproblem with objective $J_p$
        \If{feasible}
            \State $\mathbf{J}^\star\leftarrow
            \bigl(J_1(\mathbf{x}_T^\star),\ldots,
            J_l(\mathbf{x}_T^\star)\bigr)$
            \State Add $(\mathbf{u}_T^\star,\mathbf{J}^\star)$
            to $\mathcal{C}$
        \EndIf
    \EndFor
\EndFor

\State Pareto Set $\mathcal{P}\leftarrow$ non-dominated solutions in $\mathcal{C}$
\State \Return
$
\mathbf{u}_T^\star
=
\underset{(\mathbf{u}_T,\mathbf{J})\in\mathcal{P}}{\arg\min}
\left\|\mathbf{u}_T-\mathbf{u}_T^{\mathrm{nom}}\right\|_1
$

\end{algorithmic}
\end{algorithm}


Finally, we present our proposed MPC-STL framework in Algorithm~\ref{alg:pareto_epsilon}. In the algorithm, optimization \eqref{eq:strict_mpc_stl} refers to the original problem defined in Section~\ref{sec:worst-case-mpc}. The framework first provides a formal quantification of violation rate when strict satisfaction is feasible, and then offers a repair resolution that considers tradeoffs among agent-specific objectives with weak Pareto guarantees.
One major advantage of our multi-objective formulation is that it quantifies some user-defined cost metric for individual agent, which cannot be captured by a single ego-centric relaxation objective, such as obtaining a least-violating controller that ignores consequences of a relaxation for other agents. On the other hand, objective tradeoffs provide counterfactual reasoning for alternative control decisions that is systematically justifiable. We can formally and quantitatively reason how the performance of each agent is affected if a different control action is taken.

\subsection{MILP encoding}
\label{sec:milp}

In this section, we discuss how to encode the MPC-STL problems as MILPs, as well as some assumptions made. 

\textbf{Dynamics constraints.}
We assume that state space $\mathcal{X}$ and admissible control set $\mathcal{U}$ are polytopes bounded by a set of linear inequalities, e.g., $\mathcal{X}=\{x\in\mathbb{R}^{n_x}\mid Ax\leq b\}$, which can be directly encoded as linear constraints. The robot dynamics in \eqref{eq:discrete_dynamics} is not assumed to be linear, but it is linearized here for MILP constraints.

\textbf{STL constraints.}
We adopt the method proposed in \cite{raman2014}, where a binary variable $z_t^\varphi\in{0,1}$ is introduced such that
\begin{equation}
z_t^\varphi = 1 \quad \Leftrightarrow \quad (S,t)\models\varphi.
\label{eq:milp_encoding}
\end{equation}
Let $x_t := S(t)$ denote the state at time $t$. For an affine predicate $\mu$ with $l(x_t)=a^\top x_t+b$, its truth value can be encoded using the big-$M$ formulation:
\begin{equation}
l(x_t)\geq -M(1-z_t^\mu),\qquad
l(x_t)\leq Mz_t^\mu,
\label{eq:milp_constriants}
\end{equation}
where $M>0$ is a sufficiently large constant. 
Logical operators are recursively encoded using linear constraints. 
For example, negation is encoded as
\begin{equation}
    z_t^{\neg \varphi} = 1 - z_t^\varphi,
\end{equation}
while conjunction $\psi = \bigwedge_{i=1}^{m} \varphi_i$ is encoded as
\begin{equation}
\begin{aligned}
    z_t^\psi &\leq z_t^{\varphi_i}, 
    && i = 1,\ldots,m, \\
    z_t^\psi &\geq 1-m+\sum_{i=1}^{m} z_t^{\varphi_i}.
\end{aligned}
\end{equation}
These constraints ensure that $z_t^\psi=1$ if and only if 
$z_t^{\varphi_i}=1$ for all $i=1,\ldots,m$.
Under the discrete-time interpretation, the always operator $\Box_I\varphi$ is equivalent to a finite conjunction:
\begin{equation}
z_t^{\Box_I\varphi}
=
\bigwedge_{t'\in \{a, a+1, ..., b\}} z_{t'}^\varphi,
\end{equation}

\textbf{Relaxation constraints.}
We introduce spatial relaxation at the predicate level, where
\begin{equation}
\begin{aligned}
l(x_t)+\xi_k
\geq -M(1-z_t^\mu), \quad
l(x_t)+\xi_k
\leq Mz_t^\mu
\end{aligned}
\label{eq:relaxed_predicate_milp}
\end{equation}
Thus, when $z_t^\mu=1$, the relaxed predicate
$l(x_t)\geq-\xi_k$ is enforced. Setting $\xi_k=0$ recovers the original
predicate.

%% file: revision/experiments.tex
\section{experiments}

In this section, we present two autonomous driving case studies to evaluate our proposed framework. Simulations are implemented in CARLA \cite{dosovitskiy2017carla}. MILP problems are formulated using CVXPY \cite{diamond2016cvxpy} and solved with Gurobi\cite{gurobi}. All experiments are conducted on an Ubuntu 22.04 machine with an Intel Core Ultra 7 255HX CPU and 32 GB of RAM.

\begin{figure*}[t]
    \centering
    \includegraphics[width=\textwidth]{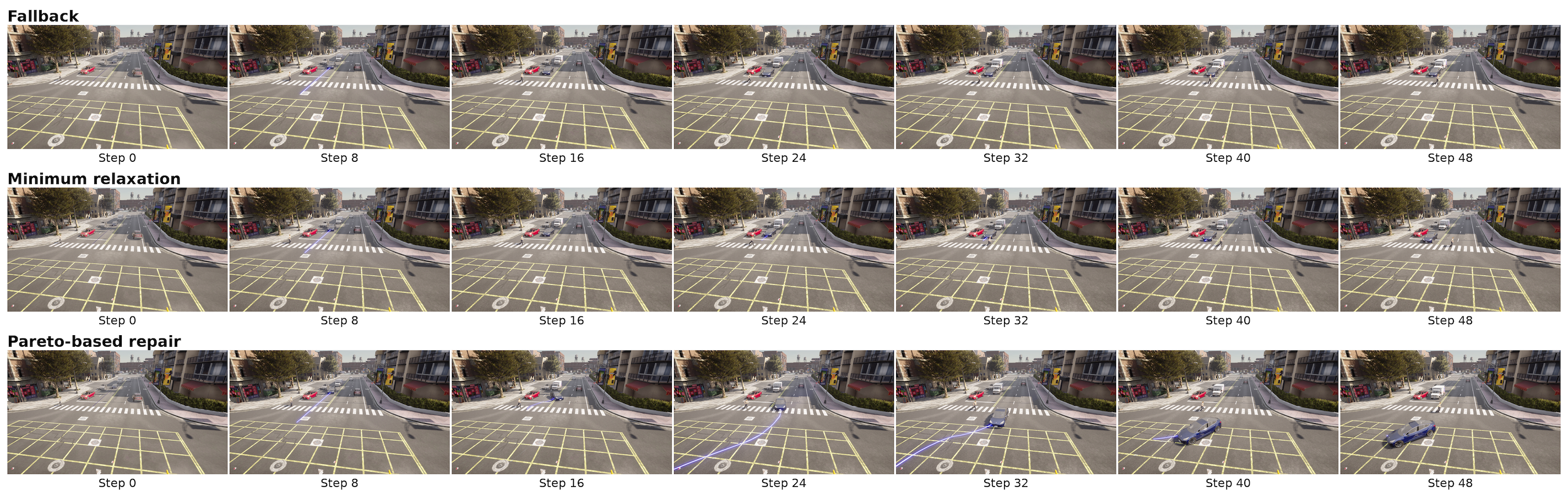}
    \caption{Exp.~1 snapshots under fallback braking, minimum relaxation, and Pareto-based lane relaxation.}
    \label{fig:exp1_simulation}
\end{figure*}

\textbf{Vehicle Model.}
We use acceleration-controlled kinematic bicycle model for ego vehicle dynamics:
\begin{equation}
\begin{bmatrix}
\dot{p}_x \\
\dot{p}_y \\
\dot{\theta} \\
\dot{v}
\end{bmatrix}
=
\begin{bmatrix}
v \cos(\theta + \beta) \\
v \sin(\theta + \beta) \\
\dfrac{v}{l_r} \sin(\beta) \\
a
\end{bmatrix},
\end{equation}
where $(p_x,p_y)$ is planar position, $\theta$ is heading, $v$ is forward speed, $a$ is longitudinal acceleration, and $\delta$ is front-wheel steering angle. The slip angle $\beta$ can be computed from $\delta$ using
\begin{equation}
\beta = \tan^{-1}\!\left( \frac{l_r}{l_f + l_r} \tan(\delta) \right),
\end{equation}
with $l_f$ and $l_r$ denoting the distances from center of mass to front and rear axles, respectively.
Assuming small slip angle ($\cos\beta \approx 1$ and $\sin\beta \approx \beta$), we use control-affine approximation \cite{thontepu2023collision}:
\begin{equation}
\underbrace{
\begin{bmatrix}
\dot{p}_x \\
\dot{p}_y \\
\dot{\theta} \\
\dot{v}
\end{bmatrix}
}_{\dot{x}}
=
\underbrace{
\begin{bmatrix}
v \cos \theta \\
v \sin \theta \\
0 \\
0
\end{bmatrix}
}_{f(x)}
+
\underbrace{
\begin{bmatrix}
0 & -v \sin \theta \\
0 & \;\, v \cos \theta \\
0 & \dfrac{v}{l_r} \\
1 & 0
\end{bmatrix}
}_{g(x)}
\underbrace{
\begin{bmatrix}
a \\
\beta
\end{bmatrix}
}_{u}.
\end{equation}
which is linearized for MILP encoding. The vehicle parameters are obtained in CARLA.

\textbf{Experiment setup.}
Each experiment is simulated for $5\,\mathrm{s}$ with planning horizon $T=2\,\mathrm{s}$ and sampling interval $dt=0.1\,\mathrm{s}$. The temporal intervals in STL specifications denote global simulation steps over which each requirement is active. At each MPC update, we enforce active specification for the planning horizon. We choose sample size $N=10$ and grid density $d=5$. Trajectory samples are generated by applying Gaussian noise to nominal control. 

\textbf{Predicates and cost functions.}
We define safe-distance predicate $\operatorname{d}(x_{ego},x_i)\geq d_{\min}$ using $\ell_\infty$-norm, where $\operatorname{d}(x_{ego},x_i) =\max\!\left\{
\left|x_{\mathrm{ego},x}-x_{i,x}\right|,
\left|x_{\mathrm{ego},y}-x_{i,y}\right|
\right\}$.
For a rectangular set
$\mathcal{R}=[r_x^{-},r_x^{+}]\times[r_y^{-},r_y^{+}]$, the state-membership predicate
$x_{\mathrm{ego}}\in\mathcal{R}$ is encoded by linear constraints
$
r_x^{-}\leq x_{\mathrm{ego},x}\leq r_x^{+},
r_y^{-}\leq x_{\mathrm{ego},y}\leq r_y^{+}.
$
Trajectory and control cost
$
J(\mathbf{x}_T,\mathbf{u}_T)
=
\left\|\mathbf{u}_T-\mathbf{u}^{\mathrm{nom}}_T\right\|_1
+
\left\|\mathbf{x}_T-\mathbf{x}^{\mathrm{nom}}_T\right\|_1.
$
Using sample-average approximation, we define
$
J_i(x_{ego}, x_i)
=
\frac{1}{N}\sum_{n=1}^{N}
\mathbb{I}\!\left\{
\exists t\in[0, T]:\operatorname{d}(x_{ego}^t, x_i^t)<d_{\mathrm{min}}
\right\}
$, which denotes the empirical collision probability with ego.

\subsection{Exp1: Intersection Pedestrian Avoidance}

\textbf{Description.} We consider a three-agent interaction with safety-critical specifications. Ego vehicle is approaching an intersection when an ambulance is coming close from behind. Ego needs to stay in its lane ($\varphi_{\text{lane}}$) and keep a safe distance from the ambulance ($\varphi_{\text{amb}}$). Next, an occluded pedestrian suddenly starts crossing the intersection and becomes visible to ego ($\varphi_{\text{ped}}$) at step 10. 

\textbf{STL specifications.}
$\varphi_{\text{lane}} =
\mathbf{G}_{[0,50]}(x_{\text{ego}}\in\mathcal{R}_{\text{lane}})$,
$\varphi_{\text{ped}} =
\mathbf{G}_{[10,50]}(d(x_{\text{ego}},x_{\text{ped}})\ge3)$,
$\varphi_{\text{amb}} =
\mathbf{G}_{[0,50]}(d(x_{\text{ego}},x_{\text{amb}})\ge5)$.

\textbf{Discussion.}
Fig.~\ref{fig:exp1_simulation} shows snapshots of our simulations in CARLA using our Pareto-based pair and two baseline methods. The fallback method applies an default brake action when having infeasible specifications. We see this abrupt brake leads to a collision with the ambulance. The minimum relaxation method recovers a control of least violation from specifications. It shows a slow brake and gradually comes to a stop, but nearly collides with the pedestrian. On the other hand, our Pareto-based repair performs a safe maneuver where ego relaxes the lane restriction and safely bypasses the pedestrian. We can see from the figure that when ego and pedestrian are parallel, there exists sufficient safe space between them. 

\begin{figure}[t]
    \centering
    \includegraphics[width=\columnwidth]{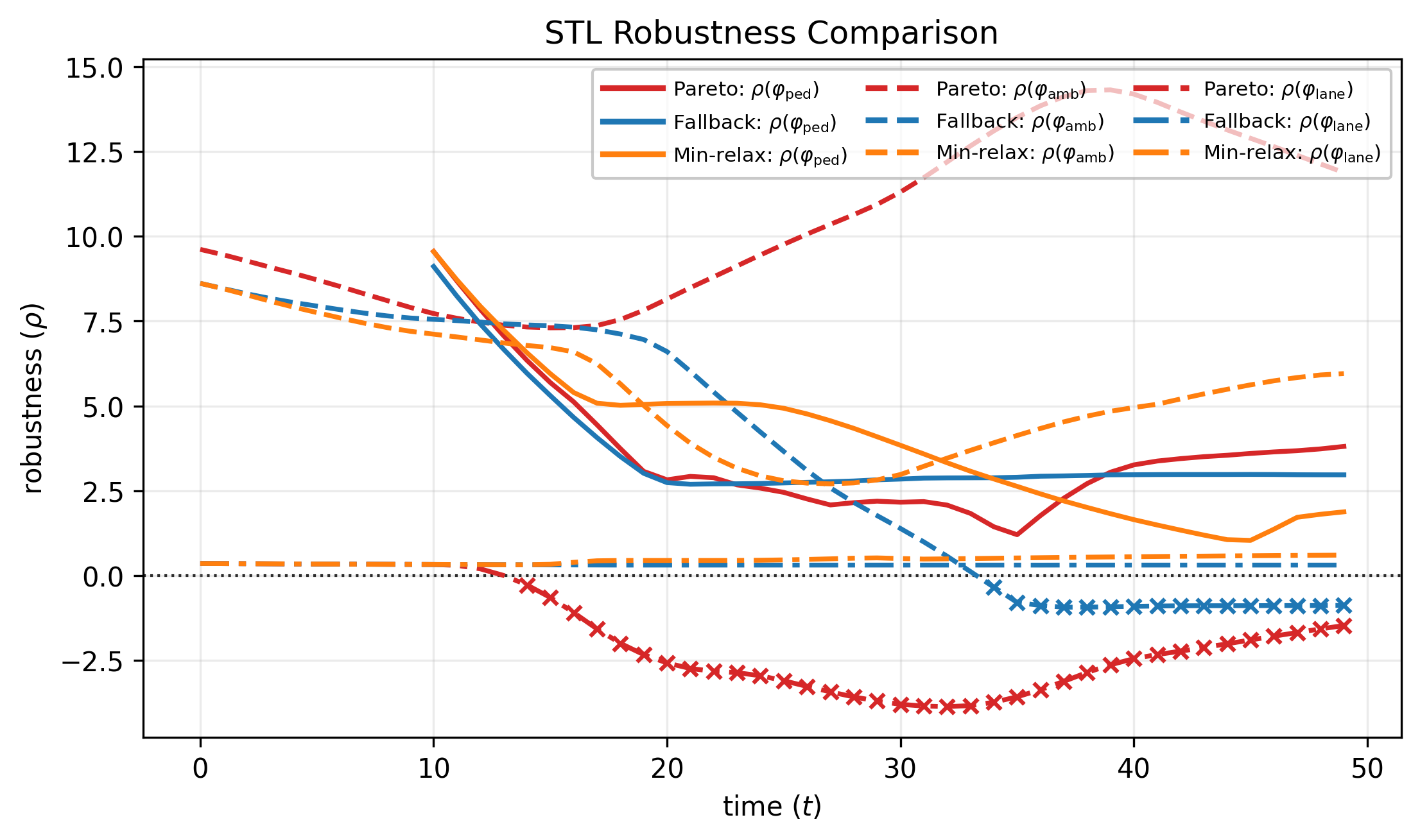}
    \caption{Comparison of STL robustness using (1) fallback controller, (2) minimum relaxation, and (3) Pareto optimization}
    \label{fig:exp1_robustness}
\end{figure}

Fig.~\ref{fig:exp1_robustness} compares robustness of specifications for each approach, which confirms our observation in Fig.~\ref{fig:exp1_simulation}. The blue dashed line for $\rho(\varphi_{amb})$ in fallback has a negative robustness starting at time 33 that reflects a collision. Our Pareto-based method relaxes $\varphi_{lane}$ at time 12, leading to a more robust satisfaction of $\varphi_{amb}$ while maintaining a $\rho(\varphi_{ped})$ comparable with baselines. This indicates that our method successfully detects solutions which can improve the safety of ambulance without sacrificing the pedestrian by relaxing the lane restriction, in accordance with our Pareto-based design.

\begin{figure}[t]
    \centering
    \includegraphics[width=\columnwidth]{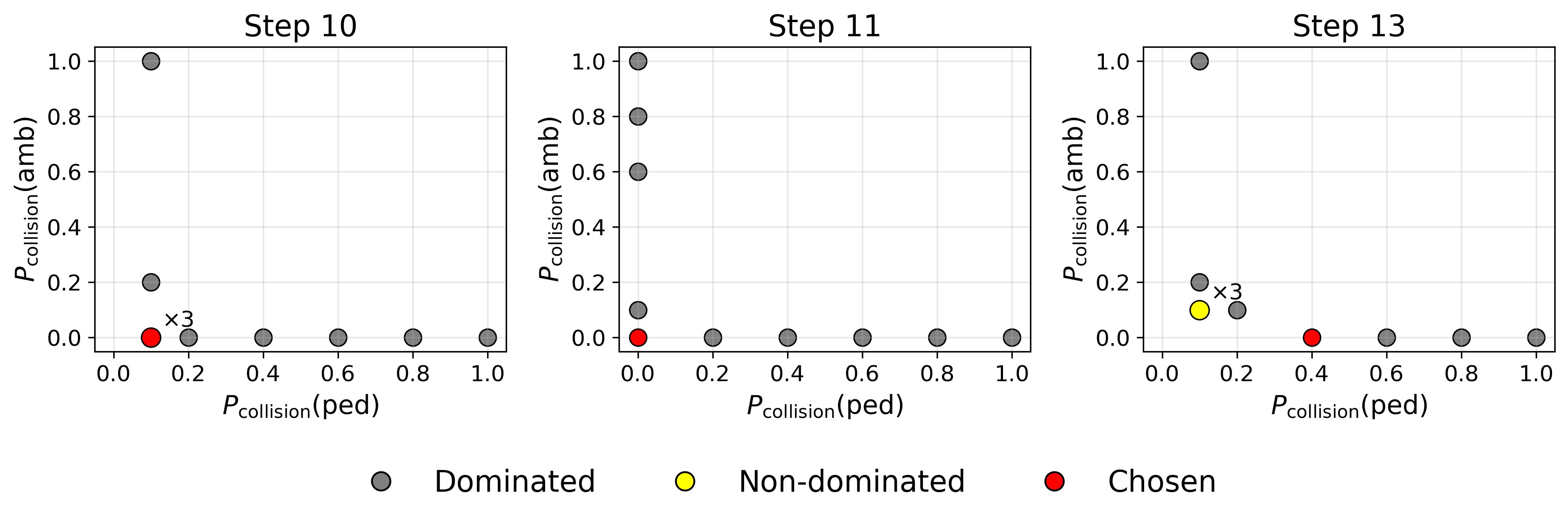}
    \caption{2D plots of discrete approximation of Pareto front in Exp1. The multiplicity number  indicates times of point repetition.}
    \label{fig:exp1_pareto}
\end{figure}

Fig.~\ref{fig:exp1_pareto} shows the computed Pareto points at three infeasible steps. Each displayed point is non-dominated within the finite candidate set generated by the selected \(\epsilon\)-grid. Because the grid provides only a discrete exploration of the feasible objective space, this observation does not establish that every retained point is Pareto optimal with respect to the original continuous feasible set.
An interesting observation is that of redundant and dominated candidate solutions. Upon further analysis, we identify two main causes. First, $\epsilon$-grids are discretized uniformly and combined through Cartesian product. Multiple threshold combinations may define similar feasible regions or produce dominated solutions. Second, the sampled risk objectives take discrete values. When multiple trajectories have same primary objective value, secondary trajectory costs, such as control smoothness, can then determine the optimizer and produce additional candidate solutions that are dominated in the risk-objective space. These candidates are removed by our filtering process in Algorithm~\ref{alg:pareto_epsilon} and hence do not affect our optimality guarantee.

We use Monte Carlo simulation to empirically assess the probabilistic statement in Theorem 1. We run the experiment 30 times with different initial locations. With $\varepsilon=0.05$ and $\beta=0.05$, Theorem~1 requires $M=59$ validation samples. At each feasible step that passes validation, we estimate $p_{\mathrm{viol}}(\mathbf{u}_T^\star)$ using 10,000 additional samples. In 250 of the 251 evaluated steps, $\hat{p}_{\mathrm{viol}}(\mathbf{u}_T^\star)<0.05$, yielding an empirical agreement rate of $0.996$. This result is consistent with Theorem~1, which guarantees $p_{\mathrm{viol}}(\mathbf{u}_T^\star)\leq 0.05$ with confidence at least $0.95$ when validation succeeds.

We report the solve times in Table~\ref{tab:computation_epsilon} computed over 30 runs. Although the simulation uses a sampling interval of \(0.1\,\mathrm{s}\), control sequences are computed offline before execution. Therefore, the reported repair times do not demonstrate real-time MPC performance. We leave for future work to develop more efficient approximation methods.

\subsection{Experiment 2: Highway Emergency Maneuver}

\textbf{Description.} Ego is driving in the rightmost lane on a highway. It is required to stay in lanes ($\varphi_{\text{highway}}$) and keep distance from a leader ($\varphi_{\text{leader}}$) and a follower vehicle ($\varphi_{\text{follower}}$). The leader performs a sudden left swerve to avoid colliding with a crash scene at step 20. In this emergency, ego needs to bypass the crash scene ($\varphi_{\text{crash}}$) while avoiding collision with a vehicle coming close in the left lane ($\varphi_{\text{left}}$) and a person walking near the scene ($\varphi_{\text{person}}$).

\textbf{STL specifications.}
$\varphi_{\text{highway}} =
\mathbf{G}_{[0,50]}(x_{\text{ego}}\in\mathcal{R}_{\text{highway}})$,
$\varphi_{\text{leader}} =
\mathbf{G}_{[0,20]}(d(x_{\text{ego}},x_{\text{leader}})\ge5)$,
$\varphi_{\text{follower}} =
\mathbf{G}_{[0,50]}(d(x_{\text{ego}},x_{\text{follower}})\ge5)$,
$\varphi_{\text{left}} =
\mathbf{G}_{[20,50]}(d(x_{\text{ego}},x_{\text{left}})\ge3)$,
$\varphi_{\text{person}} =
\mathbf{G}_{[20,50]}(d(x_{\text{ego}},x_{\text{person}})\ge2)$,
$\varphi_{\text{crash}} =
\mathbf{G}_{[20,50]}(x_{\text{ego}}\notin\mathcal{R}_{\text{crash}})$.

\begin{figure}[t]
    \centering
    \includegraphics[width=\columnwidth]{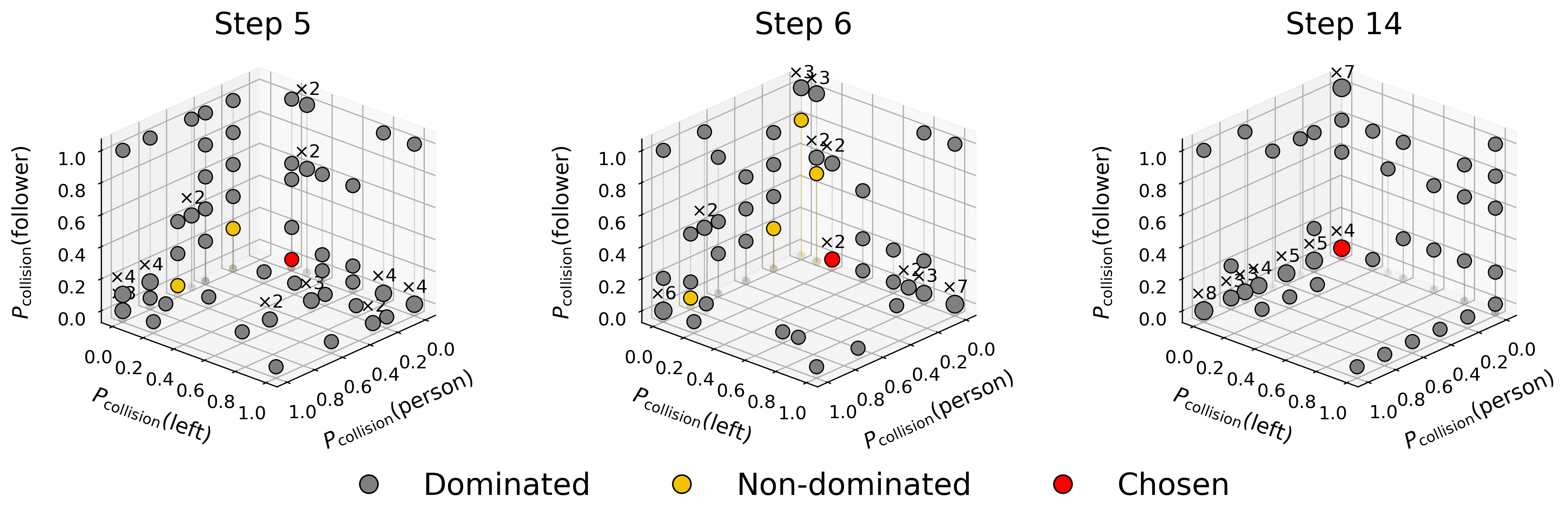}
    \caption{3D plots of discrete approximation of Pareto front in Exp2.}
    \label{fig:exp2_pareto}
\end{figure}

\textbf{Discussion.}
Fig.~\ref{fig:exp2_pareto} shows the three-dimensional Pareto points for the left vehicle, follower, and pedestrian objectives, exhibiting tradeoff patterns consistent with those observed in Exp.~1.  We also run 30 trials of Exp. 2 and report average solve time in Table~\ref{tab:computation_epsilon}. Solving one feasible MPC-STL problem takes only about $0.15\,\mathrm{s}$. However, constructing the Pareto approximation requires multiple $\varepsilon$-constraint solves, increasing the total solve time from $6.82$ to $44.8\,\mathrm{s}$ with larger density choices. With density $d=5$, each approximation requires solving 75 subproblems that results in a solve time of $44.8s$. Thus, we identify that grid enumeration is the main bottleneck for scaling to larger systems. One promising direction to resolve the exponential complexity is through adaptive grid generation that avoids enumerating the full Cartesian product. This requires refined objective ranges to reduce grid-search space and use information from previously solved subproblems to prune redundant or predictably infeasible regions.


\begin{table*}[t]
\centering
\caption{Computational Performance under Different $\epsilon$-Grid Densities. MILP size is reported per $\epsilon$-constraint subproblem. Pareto-approximation time includes all $\epsilon$-constraint solves and non-dominated filtering.}
\label{tab:computation_epsilon}
\setlength{\tabcolsep}{6pt}
\begin{tabular}{c ccc ccccc}
\toprule
Experiment
&
\multicolumn{3}{c}{Feasible Solutions}
&
\multicolumn{5}{c}{Pareto Approximation}
\\
\cmidrule(lr){2-4}
\cmidrule(lr){5-9}
&
\# cons.
&
\# vars.
&
Avg. time (s)
&
density
&
\# $\varepsilon$-problems
&
\# cons.
&
\# vars.
&
Avg. time (s)
\\
\midrule
\multirow{3}{*}{Exp.~1}
& \multirow{3}{*}{1321}
& \multirow{3}{*}{1029}
& 0.10
& $2$ & $4$
& \multirow{3}{*}{4493}
& \multirow{3}{*}{3580}
& 1.09 \\
& & & 0.10 & $3$ & $6$  & & & 1.64 \\
& & & 0.10 & $5$ & $10$ & & & 2.67 \\
\midrule
\multirow{3}{*}{Exp.~2}
& \multirow{3}{*}{2413}
& \multirow{3}{*}{1811}
& 0.15
& $2$ & $12$
& \multirow{3}{*}{6720}
& \multirow{3}{*}{5287}
& 6.82 \\
& & & 0.15 & $3$ & $27$ & & & 16.34 \\
& & & 0.15 & $5$ & $75$ & & & 44.80 \\
\bottomrule
\end{tabular}
\end{table*}

%% file: revision/conclusion.tex
\section{conclusion}

In this paper, we present a worst-case MPC framework with sampling-based STL constraints under stochastic and uncontrollable agents. We first derived a provable probabilistic certificate for feasible solutions when strict satisfaction is possible, then proposed a spatially-relaxed problem with Pareto-based feasibility repair that explicitly considers tradeoffs between agent objectives. We also show practical use of the framework through demonstrated safe behaviors in autonomous driving. The main limitation of our approach is the computational burden of solving multiple $\epsilon$-constraint subproblems at each step. Future work includes developing more efficient Pareto-front approximation by truncating $\varepsilon$-grid points online and using heuristics to construct non-uniform grid points. We expect these improvements to enable deployment of the proposed framework in time-critical systems involving more agents.